\documentclass[letterpaper, 10 pt, conference]{ieeeconf}  % Comment this line out if you need a4paper

\IEEEoverridecommandlockouts                              % This command is only needed if 
\usepackage{graphicx} % insert pdf image and \resizebox
\usepackage{bm} % to bold \mu
\usepackage{amssymb} % for real number R
\usepackage{amsmath} % for equation \text
\usepackage{multirow} % for results table multirow
\usepackage{booktabs} % for lines in table
\usepackage{xcolor}    % change grey color words
\usepackage{arydshln} % dash line for table
\usepackage{array}     % resize table column length
\usepackage[labelsep=colon]{caption}
\title{\LARGE \bf
LagMemo: Language 3D Gaussian Splatting Memory for Multi-modal Open-vocabulary Multi-goal Visual Navigation
}

\author{Haotian Zhou$^{1}$, Xiaole Wang$^{1}$, He Li$^{1}$, Jianghuan Xu$^{1}$, Zhuo Qi$^{1}$, Jinrun Yin$^{1}$, Haiyu Kong$^{2}$, Huijing Zhao$^{1}$% <-this % stops a space
\thanks{$^{1}$Haotian Zhou, Xiaole Wang, He Li, Jianghuan Xu, Zhuo Qi, Jinrun Yin, and Huijing Zhao are with the Key Laboratory of Machine Perceptions (MOE), School of Intelligence Science and Technology, Peking University, Beijing, 100084, China (e-mail: zhoutiantian19@pku.edu.cn; zhaohj@pku.edu.cn).}%
\thanks{$^{2}$Haiyu Kong is with Beijing University of Posts and Telecommunications (BUPT), Beijing, 100876, China.
}
}
\begin{document}

\maketitle
\thispagestyle{empty}
\pagestyle{empty}

%%%%%%%%%%%%%%%%%%%%%%%%%%%%%%%%%%%%%%%%%%%%%%%%%%%%%%%%%%%%%%%%%%%%%%%%%%%%%%%%
\begin{abstract}

Navigating to a designated goal using visual information is a fundamental capability for intelligent robots. To address the practical demands of multi-modal, open-vocabulary goal queries and multi-goal visual navigation, we propose LagMemo, a navigation system that leverages a language 3D Gaussian Splatting memory. During a one-time exploration, LagMemo constructs a unified 3D language memory with robust spatial-semantic correlations. With incoming task goals, the system efficiently queries the memory, predicts candidate goal locations, and integrates a local perception-based verification mechanism to dynamically match and validate goals. For fair and rigorous evaluation, we curate GOAT-Core, a high-quality core split distilled from GOAT-Bench. Experimental results show that LagMemo's memory module enables effective multi-modal open-vocabulary localization, and significantly outperforms state-of-the-art methods in multi-goal visual navigation. Project page: https://weekgoodday.github.io/lagmemo

\end{abstract}

%%%%%%%%%%%%%%%%%%%%%%%%%%%%%%%%%%%%%%%%%%%%%%%%%%%%%%%%%%%%%%%%%%%%%%%%%%%%%%%%
\section{INTRODUCTION}

\begin{figure}[t]
	\centering
	\includegraphics[scale=0.465]{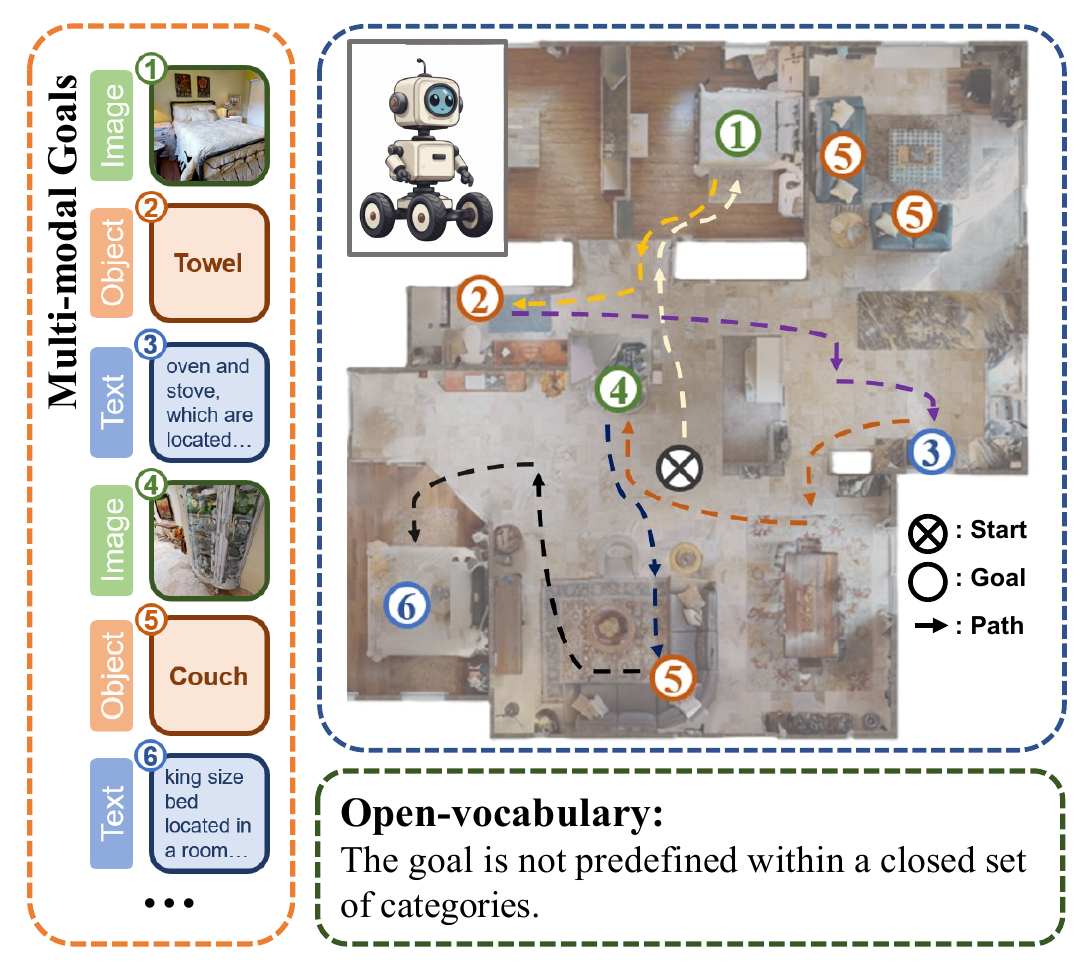}
	\vspace{-7mm}
	\caption{Illustration of multi-modal open-vocabulary multi-goal visual navigation task. Multi-modal: the goal can be specified in the forms of an object category, an image or a text description; Open-vocabulary: the agent is not limited to navigating to a predefined closed set of categories; Multi-goal: the agent is required to find multiple goals within the same environment.
}
	\label{fig:1}
	\vspace{-7mm}
\end{figure}

In real-world applications such as home assistants and service robots, mobile agents are expected to understand user instructions, perceive environments, and navigate to target objects \cite{deitke2022retrospectives}\cite{sun2024survey}. With the advancement of vision-language models, and inspired by the fact that humans primarily rely on vision to navigate, visual navigation has emerged as a prominent research area \cite{deitke2022retrospectives}\cite{wong2025survey}. This task requires robots to rely primarily on visual sensors to reach target destinations in a safe and efficient manner.
% \cite{radford2021learning}\cite{li2023blip}
Recently, the embodied navigation community has increasingly focused on the more complex and practical setting of multi-modal open-vocabulary multi-goal visual navigation, as illustrated in Fig. \ref{fig:1}. In such scenarios, robots are required to continuously complete multiple tasks within the same environment, with goals specified in various modalities and potentially involving novel categories. To advance this paradigm, GOAT-Bench \cite{khanna2024goat} has provided baseline algorithms and a dataset based on the Habitat simulator \cite{savva2019habitat}.

However, existing methods remain limited. End-to-end approaches like RL GOAT \cite{khanna2024goat} encode the environment implicitly through hidden states, leading to poor generalization. 
More importantly, while modular approaches like Modular GOAT construct an instance memory to support multi-modal queries, their memory population is fundamentally constrained by an upstream object detector with a predefined category list. From a memory perspective, this prevents open-vocabulary capability, as unforeseen novel targets are ignored during exploration and become irretrievable. For instance, an unforeseen open-set target like a ``Mickey Mouse doll'' would likely be ignored by the fixed-vocabulary detector during exploration, meaning it is never memorized and thus impossible to retrieve. Furthermore, projecting these features onto 2D semantic maps inevitably loses fine-grained 3D spatial details and lacks 3D spatial-semantic correlation, which prevents global semantic consistency optimization across multiple views. To overcome this upstream detection bottleneck and the limitations of 2D representations, we introduce LagMemo, a novel visual navigation system that integrates 3DGS with a quantized language feature space. During navigation, it supports multi-modal queries to identify candidate waypoints, improving navigation performance.

% A natural strategy for such a task is to first explore the environment to build a memory, then leverage this memory to guide task-specific searches by prioritizing most relevent regions. This strategy calls for an explicit, semantically rich, and queryable memory module that can support navigation towards multi-modal, open-vocabulary, and multi-goal targets.

\begin{figure*}[htbp]
	\centering
	\includegraphics[scale=0.373]{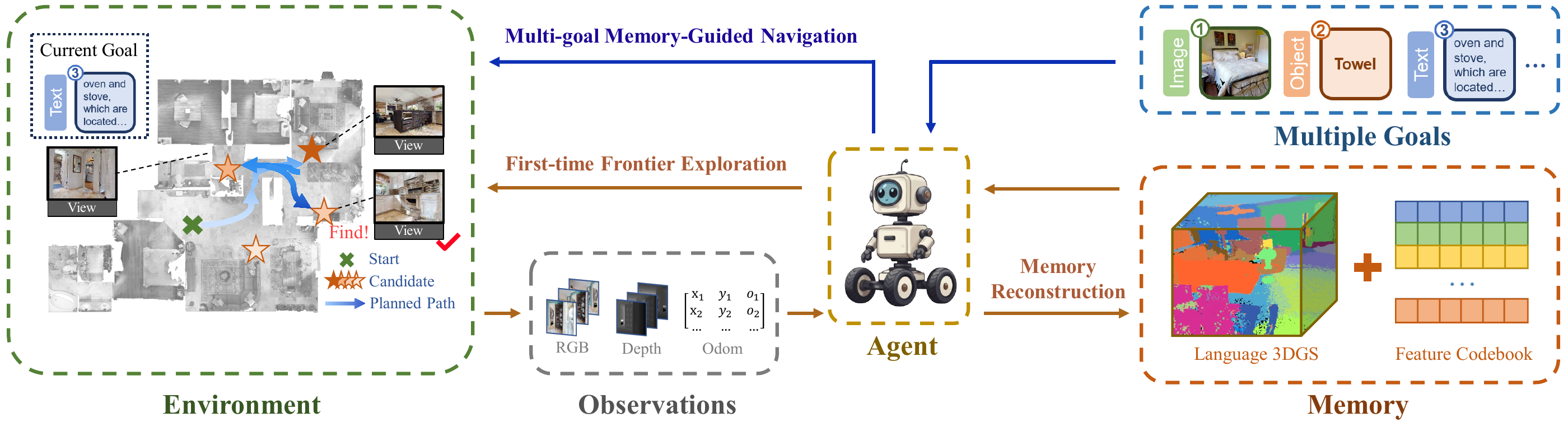}
	\vspace{-6mm}
	\caption{LagMemo Overview. The agent first performs frontier-based exploration to collect observations from the environment, upon which it reconstructs a language 3DGS memory and a feature codebook. As multi-modal open-vocabulary goals input, the agent queries the memory to generate candidate localization regions and uses real-time perception to verify targets, thereby accomplishing multi-goal visual navigation.}
	\label{fig:2}
	\vspace{-7mm}
\end{figure*}

Our main contributions are summarized as follows:
\begin{itemize}
\item We propose LagMemo, a visual navigation system that introduces a unified 3D Gaussian Splatting memory module equipped with codebook-based language feature embeddings. To address the inherently sparse observations collected during rapid pre-exploration, a keyframe retrieval mechanism is incorporated.
\item We propose a memory-guided visual navigation framework that incorporates a novel goal verification mechanism to bridge memory and real-time perception. This mechanism operates through a cyclic process of memory query and perception-based validation, effectively leveraging the constructed memory to improve navigation performance.
\item Based on GOAT-Bench, we curate a high-quality household environment split named GOAT-Core, which can be used to evaluate both goal localization and multi-modal multi-goal visual navigation tasks.
\item Extensive quantitative and qualitative evaluations, alongside real-world deployments, demonstrate that LagMemo achieves superior performance in both goal localization and multi-modal open-vocabulary multi-goal visual navigation.
\end{itemize}

\section{Related Works}

\subsection{Visual Navigation}
In terms of methodology, visual navigation approaches can be broadly categorized into modular \cite{chaplot2020object} and end-to-end \cite{ramrakhya2022habitat} frameworks, with recent advances in
zero-shot navigation methods \cite{gadre2023cows}\cite{yokoyama2024vlfm}. In terms of goal modality, visual navigation tasks are commonly classified into ObjectGoal \cite{batra2020objectnav}, ImageGoal \cite{krantz2022instance}, and TextGoal \cite{sun2024prioritized}. A critical challenge in these settings is whether the target is limited to a closed set of categories. For example, SemExp \cite{chaplot2020object} uses $C$ channels in 2D semantic map to record the location of $C$ predefined goal categories. Recent works \cite{yokoyama2024vlfm}\cite{huang2023visual} exploit CLIP or BLIP-2 to enable open-vocabulary identification and localization for novel goals. Moving beyond single-target scenarios, this paper focuses on multi-modal open-vocabulary multi-goal visual navigation, a task  formalized as lifelong visual navigation in GOAT \cite{khanna2024goat}, which is more aligned with practical applications.
%\cite{wu2024voronav}\cite{long2024instructnav}
\subsection{Memory in Visual Navigation}
Effective memory is critical for long-horizon navigation. While some approaches encode implicit memory using RNNs \cite{ramrakhya2022habitat}, explicit representations are preferred for complex environments. A common paradigm is to use 2D semantic grid maps \cite{chaplot2020object}\cite{yokoyama2024vlfm}, which discretize the environment into spatial grids, assigning each grid a semantic label or high-dimensional embedding. However, such 2D projections inevitably lose vertical spatial details and are highly susceptible to multi-view feature inconsistencies. Some works \cite{Werby2024HierarchicalO3}\cite{yin2025unigoal} construct scene graphs to model spatial relationships between objects and rooms. However, such paradigm often over-abstracts the environment by aggressively compressing dense visual information into sparse node embeddings. Recently, 3DGS has emerged as a promising dense representation (e.g., GaussNav \cite{lei2025gaussnav}, IGL-Nav \cite{Guo2025IGLNavI3}). Yet, these methods tailored to instance image-goal navigation, rely heavily on high-fidelity RGB rendering for visual matching. Such dense multi-view observations are typically not available during rapid robotic exploration. In contrast, LagMemo utilizes a codebook-quantized language 3DGS memory, preserving 3D spatial-semantic correlations while enabling efficient retrieval directly within the feature space. This design not only enables multi-modal open-vocabulary querying but also ensures robust localization even when geometric reconstruction is imperfect due to sparse exploration views.

\subsection{3D Gaussian Splatting with Language Embedding}
% 3D Gaussian Splatting (3DGS) \cite{kerbl20233d} is a 3D scene reconstruction method, which has attracted significant attention due to its high-quality and real-time rendering capabilities \cite{chen2025survey3dgaussiansplatting}. Several studies have explored online geometric modeling based on 3DGS \cite{matsuki2024gaussian}\cite{keetha2024splatam}. As many embodied tasks require not only geometric representations but also semantic guidance, recent works \cite{qin2024langsplat}\cite{zhou2024feature}\cite{shi2024language}\cite{wu2024opengaussian} have attempted to incorporate language features extracted by 2D vision-language models into 3DGS model. For example, LangSplat \cite{qin2024langsplat} compresses high-dimensional CLIP features into low-dimensional Gaussian features. Most methods \cite{qin2024langsplat}\cite{zhou2024feature}\cite{shi2024language} optimize semantic features only in the rendered 2D view, their performance degrades when directly querying semantic features on 3D Gaussians. OpenGaussian \cite{wu2024opengaussian} addresses this by proposing a feature discretization mechanism that aligns low-dimensional Gaussian features with high-dimensional 2D instance embeddings.
3D Gaussian Splatting (3DGS) \cite{kerbl20233d} is a 3D scene reconstruction method, which has attracted significant attention due to its high-quality and real-time rendering capabilities \cite{chen2025survey3dgaussiansplatting}. Beyond online geometric modeling \cite{matsuki2024gaussian}\cite{keetha2024splatam}, as many embodied tasks require not only geometric representations but also scene understanding, recent works \cite{zhou2024feature}\cite{shi2024language} attempt to incorporate language features by embedding vision-language features into Gaussians for scene understanding. For instance, LangSplat \cite{qin2024langsplat} compresses CLIP features via scene-specific autoencoders. Online Language Splatting \cite{Katragadda2025OnlineLS} enables near real-time incremental language mapping. However, while achieving high-fidelity 2D semantic rendering, these approaches exhibit limited capabilities in explicit 3D spatial indexing as pointed out in \cite{wu2024opengaussian}. LagMemo employs codebook clustering for robust 3D spatial-semantic association, and specifically tailored to the sparse-view navigation setting with efficient feature retrieval.

\section{Methodology}

\begin{figure*}[htbp]
	\centering
	\includegraphics[scale=0.475]{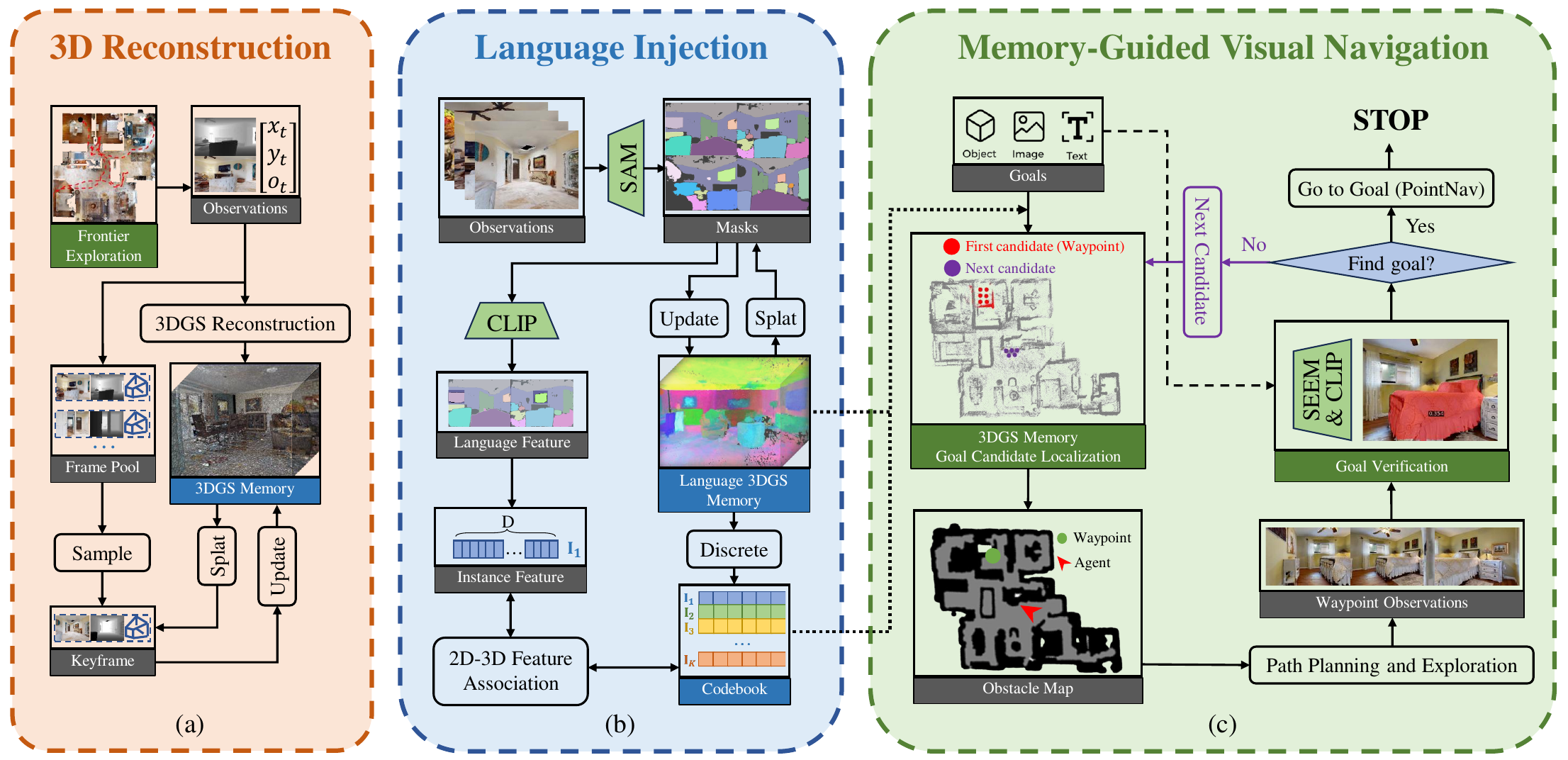}
	\vspace{-6mm}
	\caption{Language 3DGS Memory Reconstruction and Memory-Guided Visual Navigation Pipeline. (a) 3D Reconstruction. During frontier exploration, the agent collects RGB, depth, and odometry to reconstruct 3DGS memory. A keyframe retrieval mechanism is employed to mitigate the forgetting and surface holes caused by sparse navigation views. (b) Language Injection. For image observations, we leverage SAM and CLIP to extract 2D semantic features. Via 2D-3D association, these features are assigned to Gaussians and discretized into a codebook. (c) Memory-Guided Visual Navigation. During execution, multi-modal open-vocabulary goals query the memory to propose candidate locations (waypoints). Using the obstacle map for path planning, the agent verifies the target to decide success or move to the next candidate.}
	\label{fig:3}
	\vspace{-8mm}
\end{figure*}

\subsection{Task and System Overview}
\label{Sec:3.1}
We address the multi-modal, open-vocabulary, multi-goal visual navigation task in a multi-room indoor household environment. In a practical deployment scenario, an agent is expected to execute a continuous sequence of tasks within a given environment where the goal of the $k^{th}$ task is denoted as $g^k$. 
At every timestep $t$ during the task execution, the agent receives observations including an RGB image $I_t^k$, a depth image $D_t^k$, and odometry $p_t^k=(x_t^k,y_t^k,o_t^k)$, and outputs an action $A_t^k$. Upon finishing the current task, a new target goal $g^{k+1}$ is sequentially provided.
% In this task setting, every episode comprises a sequence of subtasks, where the goal of the $k^{th}$ subtask is denoted as $g^k$. At the beginning of each episode, the agent is randomly initialized. Every timestep $t$, the agent receives observations including an RGB image $I_t^k$, a depth image $D_t^k$, odometry $p_t^k=(x_t^k,y_t^k,o_t^k)$, and outputs an action $A_t^k$. Only when the current subtask is finished, will the next subtask goal $g^{k+1}$ be provided. A subtask succeeds if the agent executes STOP within a certain distance threshold from the target before exceeding the step limit. 
The navigation goals are specified in one of three modalities: a category, an image, or a text description. Crucially, the task is open-vocabulary: the agent is not limited to navigating to objects from a predefined closed set of categories.

% To efficiently complete all subtasks in an episode, we strategically sacrifice the efficiency of the first subtask and perform thorough exploration of the environment to construct a rich multi-modal, open-vocabulary spatial memory. This memory enables the agent to quickly localize and navigate to the following goals.

As illustrated in Fig.\ref{fig:2}, we propose LagMemo. Upon entering a novel environment, the agent conducts a one-time frontier-based exploration to scan the surroundings.
%The agent first conducts frontier-based exploration to explore the environment.
During this phase, a language 3D Gaussian Splatting (3DGS) memory is constructed (Sec.\ref{sec:3.2}). Once the memory is built, it serves as a persistent prior to efficiently support all subsequent multi-goal tasks within that environment.
For incoming multi-modal goals, the agent queries the language 3DGS memory, and then navigates to the candidate instances and verifies whether the observed objects match the goal (Sec.\ref{sec:3.3}).

\subsection{Language 3DGS Memory Reconstruction} 
\label{sec:3.2}

% \noindent\textbf{Frontier Exploration.} \hspace{1em}
%  The agent first performs frontier-based exploration \cite{yamauchi1998frontier} to traverse the environment. \rev{This initial phase aims for a fast, coarse scan rather than a meticulous dense traversal, resulting in inherently sparse observation viewpoints.} During this phase, the agent collects a stream of observations, including RGB, depth, and pose information, which are used for Gaussian-based geometry reconstruction. After exploration, a global optimization over the full frame sequence incorporates language features into the memory.

\noindent\textbf{Geometry Reconstruction.} \hspace{1em}
% The agent first performs frontier-based exploration \cite{yamauchi1998frontier} to collect sparse RGB-D and pose sequences. 
During the initial one-time exploration phase, the agent performs frontier-based exploration \cite{yamauchi1998frontier} to collect sparse RGB-D and pose sequences. Following \cite{keetha2024splatam}, We reconstruct 3D Gaussians parameterized by position $\bm{\mu}\in\mathbb{R}^3$, color $\bm{c}\in\mathbb{R}^3$, radius $r$, and opacity $o$. As illustrated in Fig. \ref{fig:3} (a), given incoming RGB-D frames and camera poses, new Gaussians are inserted in under-covered regions, and all parameters are optimized by the classical geometry loss that supervises both color and depth rendering:

\vspace{-5mm}
\begin{equation}
    \mathcal{L}_{geo}=(1-\lambda)\mathcal{L}_1(\bm{C})+\lambda\mathcal{L}_{SSIM}(\bm{C})+\mu\mathcal{L}_1(D)
\end{equation}
% \vspace{-2mm}

In large-scale, multi-room navigation environments with limited inter-frame overlap, geometric reconstruction often suffers from forgetting and surface holes. To mitigate this, we introduce a keyframe retrieval mechanism. 
At each timestep, the current frame is optimized for $p_1$ iterations, followed by $p_2$ iterations on sampled historical frames (weighted negatively correlated to PSNR), thus frames with lower fidelity are more likely to be revisited.
% A frame pool stores all historical frames, which are periodically rendered and evaluated by PSNR against ground-truth. At each timestep, a two-stage optimization is performed: the current frame is optimized for $p_1$ iterations to ensure accurate reconstruction of new observations, followed by $p_2$ iterations on sampled historical frames. The sampling probability of a frame is inversely related to its PSNR, thus frames with lower fidelity are more likely to be revisited. 
% \rev{Crucially, we opt for the 3DGS representation over conventional point clouds due to its volumetric and differentiable nature. While discrete point clouds typically leave 'holes' under sparse exploration views, the scale and opacity attributes of Gaussians provide a continuous surface representation that seamlessly interpolates these unobserved regions. Furthermore, this differentiable structure is fundamentally necessary for the subsequent semantic encoding. Rather than relying on heuristic multi-view feature aggregation (e.g., averaging) that is highly susceptible to inconsistencies, 3DGS enables an 'analysis-by-synthesis' optimization. By actively back-propagating rendering losses, it inherently resolves cross-view conflicts to yield a globally consistent field.}
Crucially, we opt for 3DGS over discrete point clouds due to its continuous and differentiable nature. Geometrically, Gaussians support real-time optimization and seamlessly interpolate sparse-view holes. Semantically, although optimization introduces offline overhead, this differentiable structure enables an ``analysis-by-synthesis'' paradigm. Unlike heuristic multi-view averaging, it actively back-propagates rendering losses to resolve cross-view conflicts, yielding a globally consistent and highly extensible semantic memory.

\begin{table*}[htbp]
\centering
\scriptsize
\renewcommand{\arraystretch}{1.3}
\caption{Data Comparison between GOAT-Core and GOAT-Bench of 4 Representative Rooms.}
\begin{tabular}{lccccc}
\toprule
 & \shortstack{Avg. Subtasks  per Episode} & \shortstack{Avg. Unique Categories  per Episode} & \shortstack{Avg. Inter-Subtask  Distance (m)} & \shortstack{Total  Episodes} & \shortstack{Total Subtasks %\\ (Episodes $\times$ Avg. Subtasks)
 } \\
\midrule
% \specialrule{1pt}{1pt}{1pt}
GOAT-Core (Ours) & \textbf{20} & \textbf{13.37} & \textbf{6.89} & 24 & \textbf{480} \\
\midrule
GOAT-Bench (Original) & 7.88 & 4.82 & 5.18 & \textbf{40} & 315 \\
\bottomrule
\end{tabular}
\label{tab:dataset_comparison}
\vspace{-7mm}
\end{table*}

\noindent\textbf{Language Injection.} \hspace{1em} 
Following \cite{wu2024opengaussian}, as illustrated in Fig. \ref{fig:3} (b), we incorporate language information by optimizing the feature $\bm{f}\in\mathbb{R}^d$ of each Gaussian and discretizing it into a codebook $\mathcal{C}$ which is associated with high-dimensional 2D instance features $\bm{f}_{2d}\in\mathbb{R}^D$ extracted from image observations.

Concretely, instance masks are first obtained by SAM \cite{kirillov2023segment}, and feature splatting is applied to render per-pixel semantic features. Mask-level features are aggregated and optimized with a feature loss that encourages intra-instance consistency and inter-instance separability. This initial step explicitly encodes instance boundaries, ensuring that Gaussians belonging to the same object share similar embeddings, while different objects remain distinguishable.

To achieve stable retrieval and cross-view instance alignment, Gaussian features are further discretized via a two-level codebook quantization. A coarse partition jointly considers 3D positions and language features, while a fine partition refines categories based solely on language features. Then we conduct a 2D-3D feature association between 2D instance-level features from multiple image views and the discretized 3D Gaussian language features. Specifically, for each instance category, we render all Gaussians assigned to that category and evaluate their spatial and semantic consistency with the 2D instance masks. The instance feature is then assigned to the discrete language category of the Gaussians with the highest similarity score.

This process produces a language codebook $\mathcal{C}$ where each entry corresponds to a cluster of 3D Gaussians enriched with CLIP features. By explicitly leveraging spatial correlations to filter out multi-view noise, this discretization ensures instance-level semantic consistency, thereby establishing a robust language-conditioned 3D memory for precise goal localization.

\subsection{Memory-Guided Visual Navigation}

% As illustrated in Fig. \ref{fig:3} (c), the proposed memory-guided navigation framework integrates goal localization, waypoint planning, goal verification, and final goalpoint navigation into a unified pipeline. 

\label{sec:3.3}

\noindent\textbf{Goal Localization via Memory Query.} \hspace{1em}
% Once the Language 3DGS Memory is constructed, multi-modal goals can be localized. 
The input goal, provided as either text or an image, is encoded into an embedding vector through the corresponding CLIP encoder. Cosine similarity is then computed against features of all entries stored in the codebook to identify candidate instances. The codebook records the indices of all Gaussians associated with the selected instance. Since navigation essentially requires guiding the agent towards the spatial location of the target instance, we compute the geometric centroid of all Gaussians linked to the instance and project it onto the 2D obstacle map. The resulting projection serves as the candidate position (waypoint) for navigation. %Notably, this approach is not limited by a predefined category set. By leveraging the well-aligned image-text embedding space of CLIP, it naturally supports open-vocabulary querying.

\noindent\textbf{Waypoint Navigation.} \hspace{1em}
% When the navigation system receives a new multi-modal goal, it first performs goal localization to identify the target instance and its corresponding position on the obstacle map.
After calculating the corresponding goal position on the obstacle map, the position is dilated and intersected with the traversable region to define a feasible waypoint. A collision-free path from the agent’s current position to the waypoint is then planned using the classical Fast Marching Method (FMM) \cite{sethian1996fast}.

\noindent\textbf{Goal Verification and Matching.} \hspace{1em}
% Due to incomplete observations and memory uncertainty, analogous to the blurred recollections humans may form after briefly scanning an environment, the target may not reside in the initially localized waypoint region. This necessitates a goal verification mechanism during navigation to judge whether to stop at current waypoint. 
A straightforward approach to goal verification would be applying perception models to historical frames or images rendered from the 3DGS memory before navigation. However, searching through raw frames is computationally inefficient, and more importantly, images rendered from 3DGS often contain blurring or artifacts due to the sparse views collected during rapid exploration. These artifacts severely degrade the performance of 2D matching models. Therefore, LagMemo adopts an on-site verification strategy. The 3DGS memory serves as a coarse global prior to generate candidate waypoints. 
Upon reaching a waypoint, the agent first performs a panoramic scan for goal verification. For text and object goals, we incorporate SEEM \cite{zou2023segment} model for open-vocabulary instance segmentation, generating pixel-level masks for secondary validation. A waypoint is deemed to contain the target object if either the CLIP similarity or the mask confidence exceeds the corresponding thresholds. For image goals, LightGlue \cite{lindenberger2023lightglue} feature matching is employed.  
% we observed that relying solely on CLIP similarity often leads to missed detections. Therefore, 
 If the goal is found, the system transitions into the Goalpoint Navigation stage. Otherwise, the waypoint is marked as invalid, and the memory is re-queried to obtain the next candidate instance. This iterative cycle of waypoint generation, navigation, and goal verification is repeated until the goal is found or the maximum step limit is reached.

\noindent\textbf{Goalpoint Navigation.} \hspace{1em}
Once the verification mechanism confirms the target's visibility, the system switches from the coarse memory-guided waypoint to a perception-guided final goalpoint. Using the pixel-level mask provided by SEEM in combination with current depth information, the target object's footprint is projected onto the 2D obstacle map. We define the optimal final goalpoint as the nearest traversable grid cell adjacent to this projected footprint. The agent then executes a local FMM algorithm to navigate to final goal and triggers STOP.

\section{Benchmark}
\label{sec:4}
% Our experiments are conducted on GOAT-Bench, a widely used multimodal, multi-task visual navigation benchmark. However, the original validation set exhibits several limitations in task design and data quality, which may obscure the actual performance of navigation algorithms. In particular, it provides insufficient evaluation of long-term navigation: each episode contains only about 8 subtasks on average, with high goal redundancy and short inter-goal distances, making it inadequate for assessing agents’ memory and planning in extended, continuous tasks. Moreover, issues such as inaccurate language descriptions, semantic ambiguities in object targets, missing mesh models, and cross-floor tasks further compromise fairness by causing failures attributable to environmental artifacts rather than algorithmic deficiencies.

The GOAT-Bench \cite{khanna2024goat} dataset, built on Habitat, has been the primary benchmark for evaluating multi-modal, open-vocabulary, multi-goal visual navigation. Despite its diverse scenes, several limitations hinder its ability to reasonably assess navigation capabilities. First, the evaluation of long-term memory and planning is insufficient: each episode contains only about 8 subtasks on average, with high goal repetition and relatively short inter-goal distances. Second, annotation quality issues are present: some text descriptions are inaccurate, object categories are semantically ambiguous, and certain scenes include missing meshes. These factors often cause agents to fail for reasons unrelated to algorithms, thereby obscuring fair comparison of navigation performance.

% To address these issues, we construct a high-quality core subset of the original validation set, termed GOAT-Core, designed for more rigorous evaluation of multi-modal, open-vocabulary, multi-goal navigation. GOAT-Core supports both goal localization and multi-goal visual navigation, while imposing stronger requirements on long-term memory and planning. As illustrated in Tab. \ref{tab:dataset_comparison}, we increase the number of subtasks per episode to 20 (compared to 7.88 in the original set) and enhance task diversity. The average distance between subtasks is also extended, thereby increasing the difficulty of long-range planning. Meanwhile, we select scenes with the highest official reviewer ratings and restrict all subtasks to a single floor, eliminating confounds from scene quality and cross-floor movement. Ambiguous or erroneous language descriptions are manually corrected, and semantically clear objects are prioritized as targets to guarantee annotation accuracy. Through these improvements, GOAT-Core provides a more challenging and equitable benchmark for evaluating multimodal, open-vocabulary, and multi-task navigation algorithms.

Instead of reporting on the entire GOAT-Bench validation split, which is broad but contains quality issues, we reorganize the benchmark by sampling a quality-controlled core subset GOAT-Core. As illustrated in Tab. \ref{tab:dataset_comparison}, we increase the number of subtasks per episode to 20 (compared to 7.88 in the original set) and enhance task diversity. The average distance between subtasks is also extended, imposing greater demands on memory and long-horizon planning. To ensure label reliability, we manually correct inaccurate or ambiguous text descriptions and prioritize objects with clear semantics. We further restrict all subtasks to occur on a single floor and select four reviewer-recommended high-quality multi-room scenes (\texttt{5cd}, \texttt{4ok}, \texttt{Nfv}, \texttt{Tee}), each averaging 7.25 rooms, mitigating mesh defects.

In total, GOAT-Core contains 480 multi-modal subtasks, covering 163 images, 158 objects, and 159 text goals. Curated from GOAT-Bench's ``seen", ``unseen'', and ``seen-synonyms'' validation splits, it ensures open-vocabulary evaluation.
The dataset also explicitly emphasizes long-horizon multi-goal navigation, with longer episodes and greater goal diversity.

Beyond multi-goal visual navigation, GOAT-Core also enables the evaluation of goal localization tasks. For each of the four multi-room scenes (average area 218.63 $m^2$), we conduct full-house exploration via frontier exploration, collecting RGB, depth, and odometry data. The curated 480 multi-modal goal queries with ground-truth positions can thus be used to evaluate the accuracy of goal localization after mapping (Sec. \ref{sec:5.1}).

For navigation experiments, we use the HelloRobot Stretch model as the agent. The robot has a height of 1.41~m and a base radius of 17~cm. It is equipped with an RGB camera with a resolution of $640 \times 480$, a horizontal field of view (HFOV) of $42^{\circ}$, mounted at a height of 1.31~m. The robot's depth perception range is $0.5 \sim 5.0$~m. Its action space consists of: \texttt{MOVE\_FORWARD} ($0.25$~m), \texttt{TURN\_LEFT/RIGHT} ($30^{\circ}$), \texttt{LOOK\_UP/DOWN} ($30^{\circ}$), and \texttt{STOP}. A subtask is considered successful if the agent executes the \texttt{STOP} action within a $200$-step limit and its final position is less than $1.0$~m from the target object instance.

\section{Experiments}

\subsection{Goal Localization}
\label{sec:5.1}
\noindent\textbf{Settings.} \hspace{1em} \textbf{1) Task: } The goal localization task requires mapping an open-vocabulary query to the memory constructed after frontier exploration and estimating the corresponding object’s location, evaluated on the GOAT-Core. \textbf{2) Baseline: } We use \textbf{VLMaps} \cite{huang2023visual} as baseline, which constructs a visual-language map by first back-projecting depth into a 3D point cloud to aggregate dense VLM features, and then projecting them onto a 2D grid. %Open-vocabulary goals are localized by similarity matching, producing a heatmap for path planning. It represents the state-of-the-art in 2D semantic-map navigation, though compressing 3D into 2D may lose spatial context. 
\textbf{3) Metrics: } Localization accuracy is measured by the Euclidean distance between predicted and ground-truth positions, with success defined as any of the top-5 predictions falling within a 1.5~m radius of the target. \textbf{4) Hyperparameters: } Geometry optimization runs $p_1=30$ iterations per step for new viewpoints and $p_2=60$ iterations for keyframe viewpoints. Feature dimension $d$ of Gaussians is 6. The two-level codebook uses coarse $k_1=32$ clusters and fine $k_2=5$ clusters per coarse cluster (total 160 entries).

\begin{table}[b]
\vspace{-4mm}
\centering
\caption{Goal Localization Results on GOAT-Core (across all scenes).}
\label{tab:goal_loc}
\scriptsize
\setlength{\tabcolsep}{10pt}
% \resizebox{\columnwidth}{!}{
\begin{tabular}{c c c c c}
\toprule
Method & Average & Object & Image & Text \\
\midrule
VLMaps~\cite{huang2023visual}  & 58.8\% & 69.7\% & 43.3\% & 61.0\% \\
LagMemo (Ours) & \textbf{70.8\%} & \textbf{88.4\%} & \textbf{56.4\%} & \textbf{66.8\%} \\
\bottomrule
\end{tabular}
% }
% \vspace{-6mm}
\end{table}

\noindent\textbf{Results.} \hspace{1em} In goal localization, as illustrated in Tab. \ref{tab:goal_loc}, LagMemo achieves an overall 70.8\% success rate, significantly outperforming VLMaps (58.8\%) and across all modalities. These results highlight the superiority of the globally optimized 3DGS representation over discrete point cloud aggregation for precise goal localization in complex environments.

\begin{figure}[t]
	\centering
	\includegraphics[scale=0.3]{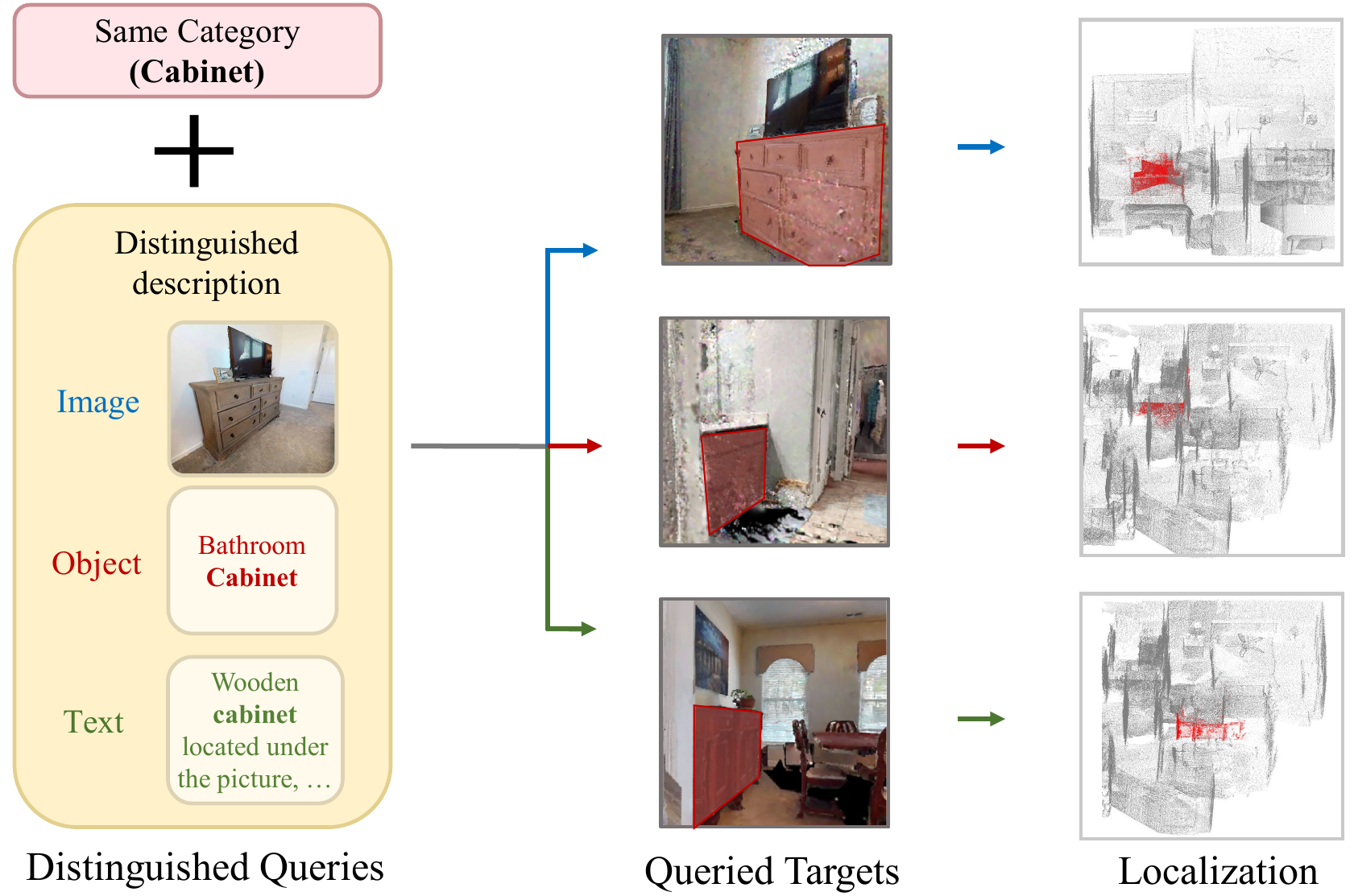}
	\vspace{-5mm}
	\caption{Distinguished Queries Retrieving Different Instances of the Same Category in Language 3DGS Memory. For the same ``cabinet'' category, with distinguished queries, the language memory can retrieve the intended target. The middle column shows a geometric rendering containing queried target, and the right column presents the 3D localization of that instance.}
	\label{fig:4}
	\vspace{-7mm}
\end{figure}

% \begin{figure}[htbp]
% 	\centering
% 	\includegraphics[scale=0.28]{fig5.pdf}
% 	\vspace{-4mm}
% 	\caption{Impact of Geometric Quality on Localization Precision. (Top) A poorly reconstructed geometric structure leads to diffuse and inaccurate localization. (Bottom) A high-quality geometry provides a strong anchor for semantic features, enabling precise localization.}
% 	\label{fig:5}
% 	\vspace{-5mm}
% \end{figure}

We also provide visualizations to demonstrate our method's precise and context-aware goal localization. As shown in Fig. \ref{fig:4}, our system is capable of fine-grained discrimination among multiple instances of the same category. 
%Fig. \ref{fig:5} illustrates that a well-reconstructed geometric structure is crucial for accurate localization, whereas poor geometry leads to diffuse and inaccurate results.
% We also provide visualizations to demonstrate our system’s target localization capabilities. Fig. \ref{fig:7-1}  shows that red-highlighted Gaussian splats match the semantics of text queries, with rendered views confirming accuracy. Fig. \ref{fig:7-2}  illustrates the importance of geometric reconstruction quality: poorly optimized 3DGS geometry leads to diffuse or incorrect localization, while high-quality geometry anchors semantic features for precise target localization, confirming that accurate 3D structure is essential for semantic injection. Fig. \ref{fig:7-3}  analyzes fine-grained discrimination among multiple instances of the same category. For complex commands like “the wooden cabinet with a mirror to the left of the painting,” CLIP encodes detailed textual features to match the correct instance. Even simpler queries such as “bathroom cabinet” are localized accurately because instance features include both appearance and contextual information from the environment. Overall, these visualizations confirm that our approach effectively integrates semantic understanding with high-quality 3D geometric reconstruction to achieve precise and context-aware target localization.

\begin{figure*}[thbp]
	\centering
	\includegraphics[scale=0.53]{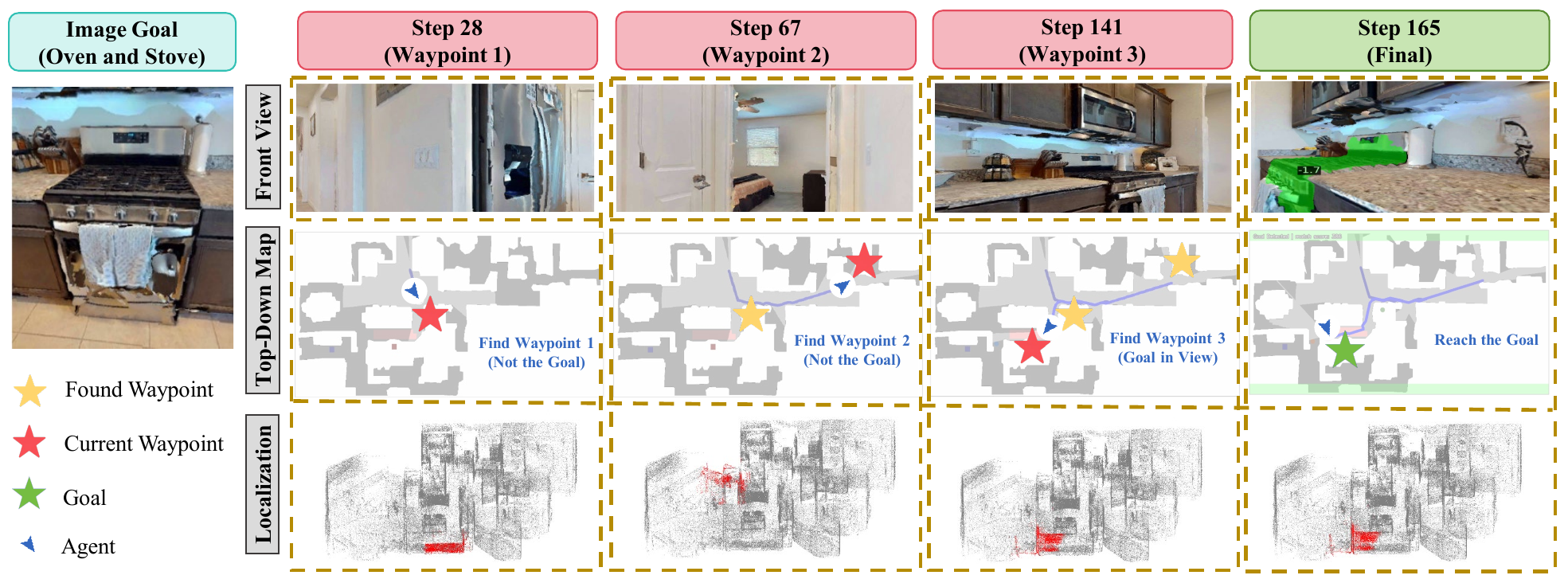}
	\vspace{-2mm}
        \caption{Step-by-step Visualization of Memory-Guided Navigation to an Image Goal. Columns show key steps (28, 67, 141, 165), rows show the front view, the top-down map, and the 3D localization results (red). In this case, the agent reaches waypoint-1/2/3 (yellow star; current waypoint in red). After checking the first two, it arrives at the third where the goal verification module identifies the goal. Then the agent proceeds to the final goal (green star) and the subtask successfully terminates at step 165.}
        
	\label{fig:7}
	\vspace{-6mm}
\end{figure*}

% 导航实验表，压缩版
\begin{table}[b]
\vspace{-4mm}
\centering
\scriptsize
\setlength{\tabcolsep}{6pt}
\caption{Multi-modal Visual Navigation Results on GOAT-Core.}
\label{tab:nav_results}

\begin{tabular}{c c c c c c}
\toprule
 & \multicolumn{2}{c}{\textsc{Overall}} & \multicolumn{3}{c}{\textsc{SR by Modality}} \\
\cmidrule(lr){2-3} \cmidrule(lr){4-6}
Method & SR ($\uparrow$) & SPL ($\uparrow$) & Object & Image & Text \\
\midrule
\textcolor{gray}{GOAT-GT Sem*} & \textcolor{gray}{75.0\%} & \textcolor{gray}{60.2\%} & \textcolor{gray}{86.4\%} & \textcolor{gray}{68.8\%} & \textcolor{gray}{76.9\%} \\
\midrule
Modular GOAT \cite{chang2023goatthing} & 38.3\% & 29.7\% & 36.6\% & 40.8\% & 37.8\% \\
GOAT Full Exp*    & 36.3\% & 28.5\% & 39.0\% & 39.5\% & 30.5\% \\
RL GOAT \cite{khanna2024goat}       & 11.3\% &  6.2\% & 18.3\% & 5.6\% & 9.2\% \\
CoWs* \cite{gadre2023cows}        & 45.8\% & 28.6\% & 58.5\% & 43.3\% & 35.4\% \\
\midrule
LagMemo (Ours) & \textbf{56.3\%} & \textbf{35.3\%} & \textbf{68.3\%} & \textbf{46.1\%} & \textbf{53.7\%} \\
\bottomrule
\end{tabular}
% \vspace{-6mm}
\end{table}

\subsection{Visual Navigation}
\noindent\textbf{Settings.} \hspace{1em} \textbf{1) Task: }
% We evaluate all methods within the Habitat simulator using the curated GOAT-core dataset, with detailed experimental settings provided in the \textit{Method} section. 
The visual navigation evaluation follows a sequential multi-goal protocol. Specifically, each evaluation episode comprises a sequence of subtasks. At the beginning of each episode, the agent is initialized at a random pose. During an episode, only when the current subtask is finished will the next subtask goal be provided to the agent. At every timestep, given visual observations as input, the agent outputs an action to interact with the environment.
\textbf{2) Baselines: } 
We compare LagMemo against five baselines. \textbf{RL GOAT} \cite{khanna2024goat}: the official reinforcement learning baseline of GOAT-Bench, which encodes multi-modal goals with CLIP and directly maps observations and goals to actions. \textbf{Modular GOAT} \cite{chang2023goatthing}: a modular approach that builds a 2D map for navigation by relying on an upstream detector to record instance images of predefined categories. \textbf{GOAT – GT Sem*}: a variant of Modular GOAT using ground-truth semantic segmentation results from simulator, serving as an upper bound for reference. \textbf{GOAT – Full Exp*}: a Modular GOAT variant with the same frontier exploration, aligned with LagMemo for fairer comparison. \textbf{CoWs*}: adapted from the CoWs \cite{gadre2023cows} method that combines frontier exploration with CLIP-based detection, extended here with multi-modal inputs. For closed set methods, we provide the list of target categories in advance to ensure feasibility, as these methods cannot operate without predefined categories. We do not compare against recent 3DGS-based image-goal methods GaussNav\cite{lei2025gaussnav} and IGL-Nav\cite{Guo2025IGLNavI3} due to their incomplete open-source and reliance on high-fidelity RGB rendering, which are incompatible with our sparse-viewpoint setting.
% \end{itemize} 
\textbf{3) Metrics: }Performance is assessed using two common metrics: SR (Success Rate), which measures the proportion of successful subtasks %(ending within $\le 1.0$m of the goal) 
and SPL (Success weighted by Path Length), which evaluates efficiency. %by comparing the trajectory length of successful episodes to the optimal path. 
\textbf{4) Hyperparameters: } The goal verification module is triggered when the agent is within 1.2~m of the candidate waypoint. For object/text goals, goal verification passes if either threshold is met: SEEM score $\tau_{SEEM} \geq 1.1$ or Mobile-CLIP cosine similarity $\tau_{CLIP} \geq 0.23$. For image goals, LightGlue declares target presence if the inlier match ratio $r_{match} \geq 5\%$.

\begin{table}[t] % 使用 [t] 单栏排版，而不是 table*
% \vspace{-1mm}
\centering
\scriptsize % 整体缩小字体
\setlength{\tabcolsep}{2.5pt} % 极限压缩列间距，防止文字重叠
\caption{Visual Navigation Results on the Full GOAT-Bench.}
\label{tab:goat_bench_full}
\resizebox{\columnwidth}{!}{% 强制缩放至单栏宽度
\begin{tabular}{c c c c c c c}
\toprule
 & \multicolumn{2}{c}{\textsc{Seen}} & \multicolumn{2}{c}{\textsc{Synonyms}} & \multicolumn{2}{c}{\textsc{Unseen}} \\
\cmidrule(lr){2-3} \cmidrule(lr){4-5} \cmidrule(lr){6-7}
Method & SR ($\uparrow$) & SPL ($\uparrow$) & SR ($\uparrow$) & SPL ($\uparrow$) & SR ($\uparrow$) & SPL ($\uparrow$) \\
\midrule
\textcolor{gray}{GOAT-GTSem \cite{khanna2024goat}} & \textcolor{gray}{56.7\%} & \textcolor{gray}{40.3\%} & \textcolor{gray}{58.4\%} & \textcolor{gray}{43.5\%} & \textcolor{gray}{54.3\%} & \textcolor{gray}{41.0\%} \\
\midrule
Modular GOAT \cite{chang2023goatthing} & 26.3\% & 17.5\% & 33.8\% & 24.4\% & 24.9\% & 17.2\% \\
Modular CoWs \cite{gadre2023cows} & 14.8\% & 8.7\% & 18.5\% & 11.5\% & 16.1\% & 10.4\% \\
\midrule
SenseAct-NN (SC) \cite{khanna2024goat} & 29.2\% & 12.8\% & 38.2\% & 15.2\% & 29.5\% & 11.3\% \\
SenseAct-NN (Mono) \cite{khanna2024goat} & 16.8\% & 9.4\% & 18.5\% & 10.1\% & 12.3\% & 6.8\% \\
\midrule
LagMemo (Ours) & \textbf{36.8\%} & \textbf{20.7\%} & \textbf{44.8\%} & \textbf{32.1\%} & \textbf{37.9\%} & \textbf{26.3\%} \\
\bottomrule
\end{tabular}
}
\vspace{-7mm}
\end{table}

\noindent\textbf{Evaluation on GOAT-Core.} We first evaluate our method on the GOAT-Core split (Sec. \ref{sec:4}) to ensure a high-quality, statistically meaningful comparison. As shown in Tab. \ref{tab:nav_results}, LagMemo significantly outperforms all baselines in overall navigation, surpassing the second-best method (CoWs*) by 10.5\% in SR and achieving the highest SPL (35.3\%). Compared to GOAT-Full Exp under the same pre-exploration setting, LagMemo improves SR by 20.0\% and SPL by 6.8\%, confirming the crucial benefit of our 3DGS memory over 2D projections. Furthermore, the modality breakdown demonstrates LagMemo's robustness across diverse query modalities. Its notable advantage on text queries specifically highlights the superiority of our language-quantized codebook for open-vocabulary retrieval.
% On average, LagMemo outperforms the second-best baseline CoWs* by 10\% in SR, and surpasses Modular GOAT by 18\%. From a modality perspective, Fig.~\ref{fig:6} shows that LagMemo demonstrates particularly strong advantages in text-based navigation tasks, outperforming all other baselines by a clear margin. Under the same pre-exploration setting, LagMemo achieves an improvement of SR +20\% and SPL +6\% compared with Modular GOAT, confirming the benefit of richer 3D semantic representations. 
% The unique strength of our goal-guided mechanism is also evident in the comparison with CoWs* (SR +10\%, SPL +6\%).
% Overall, LagMemo achieves the best SPL (35.3\%). While slightly lower SPL is observed in \texttt{4ok} and \texttt{Nfv}, this is mainly due to the strict goal verification mechanism, which sacrifices some efficiency but ensures substantially higher success rates.

% Benefiting from its superior SR, LagMemo achieves the best overall SPL on the dataset (35.3\%). Although LagMemo exhibits slightly lower SPL in \texttt{4ok} and \texttt{Nfv} compared to some baselines, we argue that this moderate efficiency loss is acceptable given the substantial improvements in success rate. Moreover, we consider the development of more flexible goal-guided exploration strategies as a promising direction for further enhancing our framework. 
% % 导航实验结果分模态图
% \begin{figure}[htbp]
% 	\centering
% 	\includegraphics[scale=0.41]{fig6.pdf}
% 	\vspace{-1mm}
% 	\caption{Navigation Performance across 3 Modalities.}
% 	\label{fig:6}
% 	\vspace{-3mm}
% \end{figure}

\noindent\textbf{Evaluation on Full GOAT-Bench.}
While GOAT-Core evaluates long-horizon memory, we further test on the full GOAT-Bench validation set to assess broad generalization. As shown in Tab. \ref{tab:goat_bench_full}, LagMemo consistently achieves the highest SR across all three splits. These comprehensive results demonstrate that LagMemo's architectural advantages persist even under the noisy and complex original benchmark.

% \begin{table}[t]
% \centering
% \small
% \setlength{\tabcolsep}{4pt}
% \renewcommand{\arraystretch}{1.1}
% \caption{Navigation performance on the GOAT-Bench validation split.}
% \vspace{2pt}
% \begin{tabular}{lcccccc}
% \toprule
% \multirow{2}{*}{Method} & \multicolumn{2}{c}{\textbf{Val Seen}} & \multicolumn{2}{c}{\textbf{Val Seen Synonyms}} & \multicolumn{2}{c}{\textbf{Val Unseen}} \\
% & SR (\%) & SPL (\%) & SR (\%) & SPL (\%) & SR (\%) & SPL (\%) \\
% \midrule
% Modular GOAT~\cite{chang2023goatthing}         & 26.3 & \textbf{17.5} & 33.8 & \textbf{24.4} & 24.9 & \textbf{17.2} \\
% Modular CLIP on Wheels~\cite{gadre2023cows}   & 14.8 & 8.71 & 18.5 & 11.5 & 16.1 & 10.4 \\
% SenseAct Skill Chain~\cite{khanna2024goat}                 & \textbf{29.2} & 12.8 & \textbf{38.2} & 15.2 & \textbf{29.5} & 11.3 \\
% SenseAct Monolithic~\cite{khanna2024goat}                  & 16.8 & 9.4  & 18.5 & 10.1 & 12.3 & 6.8 \\
% \bottomrule
% \end{tabular}
% \vspace{-6pt}
% \label{tab:goatbench}
% \end{table}

\noindent\textbf{Qualitative Case Study.} Fig. \ref{fig:7} visualizes a step-by-step navigation task for an image goal (``oven and stove"), which demonstrates an effective interplay between long-term memory guidance and goal verification mechanism. 
%Fig. \ref{fig:8} highlights LagMemo's robustness across object, image, and text goals, where the baseline fails, which underscores the advantage of our 3DGS memory. 

\noindent\textbf{Failure Case Analysis.}
We compare LagMemo against GOAT-Full Exp across three failure types: Memory Indexing Error (failure to retrieve the correct target location from memory), Verification Error (failure to confirm the target locally), and Navigation Error (e.g., collisions or loops). Overall, compared to Modular GOAT, LagMemo reduces the total failure rate from 63.7\% to 43.7\%. Specifically, Memory Indexing Error drops significantly from 37.1\% to 22.1\%, and Verification Error decreases from 22.1\% to 18.3\%. These reductions highlight two core advantages: first, our codebook-quantized 3DGS memory effectively preserves 3D spatial context and filters multi-view noise to ensure reliable retrieval; second, our multimodal on-site verification module robustly handles local target confirmation.

\begin{table}[t]
\centering
\caption{Ablation Studies on Memory and Verification.}
\vspace{-2mm}
\label{tab:ablations}
\scriptsize

\textbf{(a) Memory Reconstruction (Goal Localization Accuracy)} \\
\vspace{0.1cm}
\resizebox{\columnwidth}{!}{
\begin{tabular}{cc|c|cccc}
\toprule
Keyframe & Codebook & PSNR & Avg. & Obj. & Img. & Text \\
\midrule
$\times$ & \checkmark & 21.15 & 66.3\% & 77.5\% & 57.5\% & 63.4\% \\
\checkmark & $\times$ & \textbf{27.20} & 34.6\% & 41.6\% & 21.0\% & 37.1\% \\
\checkmark & \checkmark & \textbf{27.20} & \textbf{70.8\%} & \textbf{88.4\%} & \textbf{56.4\%} & \textbf{66.8\%} \\
\bottomrule
\end{tabular}
}

\vspace{0.1cm}

\textbf{(b) Goal Verification (Navigation SR \& SPL)} \\
\vspace{0.1cm}
\resizebox{\columnwidth}{!}{
\begin{tabular}{cc|cc|ccc}
\toprule
\multirow{2}{*}{Image Match} & \multirow{2}{*}{Text Match} & \multicolumn{2}{c|}{Average} & Obj. & Img. & Text \\
\cmidrule{3-7}
 & & SR ($\uparrow$) & SPL ($\uparrow$) & SR ($\uparrow$) & SR ($\uparrow$) & SR ($\uparrow$) \\
\midrule
$\times$ (No Verif.) & $\times$ (No Verif.) & 41.3\% & 30.4\% & 45.1\% & 32.9\% & 45.1\% \\
CLIP & CLIP & 46.7\% & 30.3\% & 52.4\% & 43.4\% & 43.9\% \\
LightGlue & SEEM + CLIP & \textbf{56.3\%} & \textbf{35.3\%} & \textbf{68.3\%} & \textbf{46.1\%} & \textbf{53.7\%} \\
\bottomrule
\end{tabular}
}
\vspace{-7mm}
\end{table}

% \begin{figure}[htbp]
% 	\centering
% 	\includegraphics[scale=0.375]{fig8.pdf}
% 	\vspace{-3mm}
%         \caption{Qualitative Comparison of LagMemo against the Modular GOAT (Baseline). We present three cases with different goal modality: object (left), image (middle), text (right). LagMemo retrieves the intended instance from the language 3DGS memory, plans on the top-down map, and verifies success. Modular GOAT (baseline) either selects a wrong instance or fails to find any instance, reflecting the limits of its 2D semantic memory to form a complete and correct representation of the scene.}
        
% 	\label{fig:8}
% 	\vspace{-6mm}
% \end{figure}

% 实物实验
\begin{figure*}[t]
	\centering
	\includegraphics[scale=0.6]{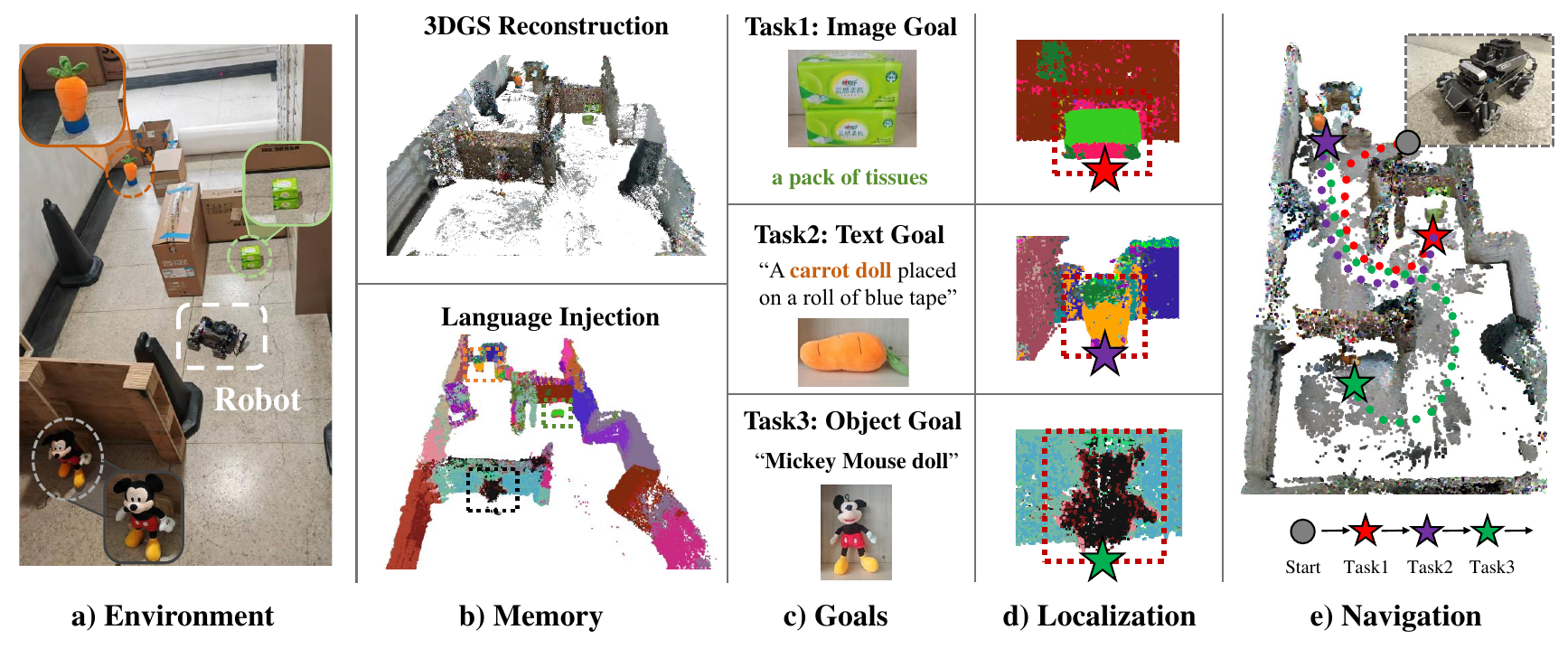}
	\vspace{-3mm}
	\caption{Real-world Deployment of LagMemo. In a physical indoor environment, we reconstruct a 3DGS memory, and successfully locate and navigate to sequential multi-modal open-vocabulary goals, such as a ``Mickey Mouse'' doll.}
	\label{fig:9}
	\vspace{-7mm}
\end{figure*}

\subsection{Ablation Study}

\noindent\textbf{Memory Reconstruction.} Tab. \ref{tab:ablations}(a) validates our memory design. Removing the keyframe mechanism degrades geometric quality and lowers localization accuracy, confirming that accurate geometry is foundational for goal localization. Furthermore, replacing the discrete codebook with a 2D-trained autoencoder drastically reduces accuracy. This highlights that explicit 3D spatial clustering is indispensable for consistent multi-room memory.

\noindent\textbf{Goal Verification.}
% In the previous subsection, we validated the accuracy of our localization strategy through ablation experiments. Once the region of interest is localized, it is necessary to determine the precise stopping point for navigation via goal verification. In this subsection, we investigate three different strategies for goal verification: (1) stopping directly at the first localized region of interest. (2) using CLIP-based similarity scoring to decide whether to stop, and (3) our final approach, which leverages LightGlue for verification in the image modality and CLIP for the text modality, combined with SEEM-based segmentation. We observe that coupling immediate goal verification with memory-guided navigation substantially improves performance. As shown in Table~\ref{tab:goal_verification}, our complete approach achieves the best overall results.
Tab. \ref{tab:ablations}(b) analyzes the verification module. Stopping without verification yields only a 41.3\% SR. A naive CLIP-based similarity marginally improves SR to 46.7\%. Our modality-specific strategy (LightGlue for images, SEEM+CLIP for text/objects) achieves the highest SR (56.3\%) and SPL (35.3\%), proving its necessity in robustly confirming targets and mitigating memory noise.
% We perform an ablation study to analyze the impact of our goal verification module, with results presented in Tab \ref{tab:goal_verification}. The baseline approach, which stops at the first reached region without any verification, achieves an average SR of 41.3\%. Introducing a generic CLIP-based similarity scoring yields a modest improvement, increasing the SR to 46.7\%. Our final approach employs the modality-specific strategy (see Sec. \ref{sec:3.3}), achieving the highest performance with an SR of 56.3\% and an SPL of 35.3\%. This demonstrates that a sophisticated, modality-specific verification mechanism is crucial for confirming the final stopping point, significantly enhancing both success rate and navigation efficiency.

\subsection{Efficiency Analysis}
We report the memory construction time, storage, and query latency on an NVIDIA RTX A6000 GPU, operating in an average 200 $m^2$ scene. As shown in Tab. \ref{tab:efficiency}(a), LagMemo requires higher offline build time and storage due to the dense 3DGS optimization. While this time is primarily dominated by comprehensive multi-view feature learning, it represents a strictly one-time offline cost per environment. By investing computational time in global spatial-semantic optimization during pre-exploration, LagMemo explicitly resolves multi-view feature inconsistencies and establishes robust 3D spatial correlations. Consequently, our discrete codebook enables fast index lookups against the established memory, achieving a near-instantaneous $0.5$s query latency.

Furthermore, Tab. \ref{tab:efficiency}(b) details the per-step inference latency during navigation. The goal verification module operates conditionally (executing specific matching models only when necessary). Combined with the planner, the total inference time is 626ms per step. This confirms that despite the mapping cost, LagMemo ensures real-time navigation, making it well-suited for deployment on real-world robotic platforms.

\begin{table}[t]
\centering
\scriptsize
\setlength{\tabcolsep}{3pt}
\caption{System Efficiency.}
\vspace{-2mm}
\label{tab:efficiency}
\begin{minipage}{\columnwidth}
\centering
\textbf{(a) Memory Building and Query Efficiency} \\
\vspace{0.1cm}
\begin{tabular}{cccc}
\toprule
Method & Build Time (s) $\downarrow$ & Query Latency (s) $\downarrow$ & Storage (MB) $\downarrow$ \\
\midrule
GOAT \cite{chang2023goatthing} & \textbf{1260} & $>$10$^*$ & $\sim$400 \\
VLMaps \cite{huang2023visual} & $\sim$2000 & 1.1 & $\sim$\textbf{200} \\
LagMemo (Ours) & $\sim$4200 & \textbf{0.5} & $\sim$500 \\
\bottomrule
\end{tabular}

\vspace{3pt}
\begin{minipage}{0.92\columnwidth}
\scriptsize\raggedright{$^*$ Matching takes 0.23s per image; GOAT stores hundreds of images, leading to high query latency.}
\end{minipage}
\end{minipage}

\vspace{0.1cm}
\begin{minipage}{\columnwidth}
\setlength{\tabcolsep}{9pt}
\centering
\textbf{(b) Per-Step Inference Time during Navigation} \\
\vspace{0.1cm}
\begin{tabular}{llc}
\toprule
Module Category & Component & Latency (ms) \\
\midrule
\multirow{3}{*}{Goal Verification} & Mobile CLIP & 225 \\
 & SEEM (Text Goal) & 190 \\
 & LightGlue (Image Goal) & 152 \\
 \cmidrule{2-3}
 & \textit{Subtotal} & \textit{396} \\
\midrule
Action Planning & FMM Planner & 230 \\
\midrule
\textbf{Total Inference Time} & \textit{per step} & \textbf{626} \\
\bottomrule
\end{tabular}

\end{minipage}
\vspace{-6mm}
\end{table}

\subsection{Real-world Deployment}

As shown in Fig. \ref{fig:9}, we deploy our system on a physical differential-drive robot. The hardware setup comprises an onboard NVIDIA Jetson Orin NX and a Realsense D435i RGB-D camera. The language 3DGS memory construction is offloaded to a remote server equipped with an NVIDIA RTX A6000 GPU, while the real-time perception, goal verification, and path planning run entirely onboard the Jetson Orin NX. Although the inaccuracy of depth camera and inherent odometry drift lead to a sub-optimal geometric reconstruction, LagMemo's codebook-quantized language memory demonstrated robustness, and successfully localizes multi-modal open-vocabulary queries (e.g., ``Mickey Mouse doll'') and navigates to the intended instances. 

\section{CONCLUSIONS}
We present LagMemo, a navigation system that integrates language 3D Gaussian Splatting to address the practical demands of visual navigation with multi-modal and open-vocabulary goal inputs as well as multi-goal tasks. By encoding language features into a codebook-quantized memory through a one-time exploration and employing an on-site perception-based goal verification mechanism, LagMemo effectively bridges global scene memory with local perception, enabling efficient goal localization and reliable navigation. Extensive experiments and real-world deployment demonstrate the superiority and practical efficiency of LagMemo.

Despite the encouraging results in integrating semantic 3D memory with visual navigation, several directions remain open. (1) Memory-aware Exploration. LagMemo’s goal localization capability hinges on geometric fidelity. Insufficient view coverage leaves blind spots in memory. Future work includes developing memory-aware active exploration that leverages geometric and semantic uncertainty to select informative views. (2) Incremental Semantics and Online Memory Construction. Currently, our global optimization and codebook clustering are performed post-exploration to ensure maximum consistency. Real-world environments, however, are dynamic, requiring adaptation. Drawing inspiration from recent advances in online language splatting, our future work will focus on transitioning to a fully incremental language 3DGS architecture, enabling both geometric optimization and online codebook growth to run concurrently during navigation. (3) Task-specific Memory Compression. Full-scene 3D representations are memory-intensive and compute-intensive. A promising direction to further improve both time efficiency and storage overhead is to pursue hierarchical or multi-resolution Gaussians, uncertainty-aware pruning, and feature compression explicitly tailored to embodied navigation needs.
%More effective active exploration strategies are needed to mitigate blind spots and ensure comprehensive memory coverage. Moreover, developing incremental memory mechanisms that enable language 3DGS to adapt to dynamic environments is a promising avenue for future research.

\addtolength{\textheight}{-12mm}   % This command serves to balance the column lengths



% \section*{ACKNOWLEDGMENT}

% \clearpage

\bibliographystyle{IEEEtran}
\bibliography{refs}

@article{deitke2022retrospectives,
  title={Retrospectives on the embodied ai workshop},
  author={Deitke, Matt and Batra, Dhruv and Bisk, Yonatan and Campari, Tommaso and Chang, Angel X and Chaplot, Devendra Singh and Chen, Changan and D'Arpino, Claudia P{\'e}rez and Ehsani, Kiana and Farhadi, Ali and others},
  journal={arXiv preprint arXiv:2210.06849},
  year={2022}
}

@article{sun2024survey,
  title={A survey of object goal navigation},
  author={Sun, Jingwen and Wu, Jing and Ji, Ze and Lai, Yu-Kun},
  journal={IEEE Transactions on Automation Science and Engineering},
  year={2024},
  publisher={IEEE}
}

@article{wong2025survey,
  title={A Survey of Robotic Navigation and Manipulation with Physics Simulators in the Era of Embodied AI},
  author={Wong, Lik Hang Kenny and Kang, Xueyang and Bai, Kaixin and Zhang, Jianwei},
  journal={arXiv preprint arXiv:2505.01458},
  year={2025}
}

@inproceedings{savva2019habitat,
  title={Habitat: A platform for embodied ai research},
  author={Savva, Manolis and Kadian, Abhishek and Maksymets, Oleksandr and Zhao, Yili and Wijmans, Erik and Jain, Bhavana and Straub, Julian and Liu, Jia and Koltun, Vladlen and Malik, Jitendra and others},
  booktitle={Proceedings of the IEEE/CVF international conference on computer vision},
  pages={9339--9347},
  year={2019}
}

@article{batra2020objectnav,
  title={Objectnav revisited: On evaluation of embodied agents navigating to objects},
  author={Batra, Dhruv and Gokaslan, Aaron and Kembhavi, Aniruddha and Maksymets, Oleksandr and Mottaghi, Roozbeh and Savva, Manolis and Toshev, Alexander and Wijmans, Erik},
  journal={arXiv preprint arXiv:2006.13171},
  year={2020}
}

@article{krantz2022instance,
  title={Instance-specific image goal navigation: Training embodied agents to find object instances},
  author={Krantz, Jacob and Lee, Stefan and Malik, Jitendra and Batra, Dhruv and Chaplot, Devendra Singh},
  journal={arXiv preprint arXiv:2211.15876},
  year={2022}
}

@inproceedings{sun2024prioritized,
  title={Prioritized semantic learning for zero-shot instance navigation},
  author={Sun, Xinyu and Liu, Lizhao and Zhi, Hongyan and Qiu, Ronghe and Liang, Junwei},
  booktitle={European Conference on Computer Vision},
  pages={161--178},
  year={2024},
  organization={Springer}
}

@article{chaplot2020object,
  title={Object goal navigation using goal-oriented semantic exploration},
  author={Chaplot, Devendra Singh and Gandhi, Dhiraj Prakashchand and Gupta, Abhinav and Salakhutdinov, Russ R},
  journal={Advances in Neural Information Processing Systems},
  volume={33},
  pages={4247--4258},
  year={2020}
}

@inproceedings{huang2023visual,
  title={Visual language maps for robot navigation},
  author={Huang, Chenguang and Mees, Oier and Zeng, Andy and Burgard, Wolfram},
  booktitle={2023 IEEE International Conference on Robotics and Automation (ICRA)},
  pages={10608--10615},
  year={2023},
  organization={IEEE}
}

@inproceedings{ramrakhya2022habitat,
  title={Habitat-web: Learning embodied object-search strategies from human demonstrations at scale},
  author={Ramrakhya, Ram and Undersander, Eric and Batra, Dhruv and Das, Abhishek},
  booktitle={Proceedings of the IEEE/CVF conference on computer vision and pattern recognition},
  pages={5173--5183},
  year={2022}
}

@inproceedings{gadre2023cows,
  title={Cows on pasture: Baselines and benchmarks for language-driven zero-shot object navigation},
  author={Gadre, Samir Yitzhak and Wortsman, Mitchell and Ilharco, Gabriel and Schmidt, Ludwig and Song, Shuran},
  booktitle={Proceedings of the IEEE/CVF Conference on Computer Vision and Pattern Recognition},
  pages={23171--23181},
  year={2023}
}

@inproceedings{yokoyama2024vlfm,
  title={Vlfm: Vision-language frontier maps for zero-shot semantic navigation},
  author={Yokoyama, Naoki and Ha, Sehoon and Batra, Dhruv and Wang, Jiuguang and Bucher, Bernadette},
  booktitle={2024 IEEE International Conference on Robotics and Automation (ICRA)},
  pages={42--48},
  year={2024},
  organization={IEEE}
}

@misc{chang2023goatthing,
      title={GOAT: GO to Any Thing}, 
      author={Matthew Chang and Theophile Gervet and Mukul Khanna and Sriram Yenamandra and Dhruv Shah and So Yeon Min and Kavit Shah and Chris Paxton and Saurabh Gupta and Dhruv Batra and Roozbeh Mottaghi and Jitendra Malik and Devendra Singh Chaplot},
      year={2023},
      eprint={2311.06430},
      archivePrefix={arXiv},
      primaryClass={cs.RO}
}

@inproceedings{khanna2024goat,
  title={Goat-bench: A benchmark for multi-modal lifelong navigation},
  author={Khanna, Mukul and Ramrakhya, Ram and Chhablani, Gunjan and Yenamandra, Sriram and Gervet, Theophile and Chang, Matthew and Kira, Zsolt and Chaplot, Devendra Singh and Batra, Dhruv and Mottaghi, Roozbeh},
  booktitle={Proceedings of the IEEE/CVF Conference on Computer Vision and Pattern Recognition},
  pages={16373--16383},
  year={2024}
}

@article{Werby2024HierarchicalO3,
  title={Hierarchical Open-Vocabulary 3D Scene Graphs for Language-Grounded Robot Navigation},
  author={Abdelrhman Werby and Chen Huang and Martin B{\"u}chner and Abhinav Valada and Wolfram Burgard},
  journal={arXiv preprint arXiv:2403.17846},
  year={2024}
}

@article{yin2025unigoal,
  title={Unigoal: Towards universal zero-shot goal-oriented navigation},
  author={Yin, Hang and Xu, Xiuwei and Zhao, Lingqing and Wang, Ziwei and Zhou, Jie and Lu, Jiwen},
  journal={arXiv preprint arXiv:2503.10630},
  year={2025}
}

@article{Guo2025IGLNavI3,
  title={IGL-Nav: Incremental 3D Gaussian Localization for Image-goal Navigation},
  author={Wenxuan Guo and Xiuwei Xu and Hang Yin and Ziwei Wang and Jianjiang Feng and Jie Zhou and Jiwen Lu},
  journal={arXiv preprint arXiv:2508.00823},
  year={2025},
}

@article{lei2025gaussnav,
  title={Gaussnav: Gaussian splatting for visual navigation},
  author={Lei, Xiaohan and Wang, Min and Zhou, Wengang and Li, Houqiang},
  journal={IEEE Transactions on Pattern Analysis and Machine Intelligence},
  year={2025},
  publisher={IEEE}
}

@inproceedings{matsuki2024gaussian,
  title={Gaussian splatting slam},
  author={Matsuki, Hidenobu and Murai, Riku and Kelly, Paul HJ and Davison, Andrew J},
  booktitle={Proceedings of the IEEE/CVF Conference on Computer Vision and Pattern Recognition},
  pages={18039--18048},
  year={2024}
}

@inproceedings{qin2024langsplat,
  title={Langsplat: 3d language gaussian splatting},
  author={Qin, Minghan and Li, Wanhua and Zhou, Jiawei and Wang, Haoqian and Pfister, Hanspeter},
  booktitle={Proceedings of the IEEE/CVF Conference on Computer Vision and Pattern Recognition},
  pages={20051--20060},
  year={2024}
}

@inproceedings{zhou2024feature,
  title={Feature 3dgs: Supercharging 3d gaussian splatting to enable distilled feature fields},
  author={Zhou, Shijie and Chang, Haoran and Jiang, Sicheng and Fan, Zhiwen and Zhu, Zehao and Xu, Dejia and Chari, Pradyumna and You, Suya and Wang, Zhangyang and Kadambi, Achuta},
  booktitle={Proceedings of the IEEE/CVF Conference on Computer Vision and Pattern Recognition},
  pages={21676--21685},
  year={2024}
}

@inproceedings{shi2024language,
  title={Language embedded 3d gaussians for open-vocabulary scene understanding},
  author={Shi, Jin-Chuan and Wang, Miao and Duan, Hao-Bin and Guan, Shao-Hua},
  booktitle={Proceedings of the IEEE/CVF Conference on Computer Vision and Pattern Recognition},
  pages={5333--5343},
  year={2024}
}

@misc{chen2025survey3dgaussiansplatting,
      title={A Survey on 3D Gaussian Splatting}, 
      author={Guikun Chen and Wenguan Wang},
      year={2025},
      eprint={2401.03890},
      archivePrefix={arXiv},
      primaryClass={cs.CV},
}

@article{Katragadda2025OnlineLS,
  title={Online Language Splatting},
  author={Saimouli Katragadda and Cho-Ying Wu and Yuliang Guo and Xinyu Huang and Guoquan Huang and Liu Ren},
  journal={arXiv preprint arXiv:2503.09447},
  year={2025}
}

@inproceedings{yamauchi1998frontier,
  title={Frontier-based exploration using multiple robots},
  author={Yamauchi, Brian},
  booktitle={Proceedings of the second international conference on Autonomous agents},
  pages={47--53},
  year={1998}
}

@inproceedings{keetha2024splatam,
  title={Splatam: Splat track \& map 3d gaussians for dense rgb-d slam},
  author={Keetha, Nikhil and Karhade, Jay and Jatavallabhula, Krishna Murthy and Yang, Gengshan and Scherer, Sebastian and Ramanan, Deva and Luiten, Jonathon},
  booktitle={Proceedings of the IEEE/CVF Conference on Computer Vision and Pattern Recognition},
  pages={21357--21366},
  year={2024}
}

@article{wu2024opengaussian,
  title={Opengaussian: Towards point-level 3d gaussian-based open vocabulary understanding},
  author={Wu, Yanmin and Meng, Jiarui and Li, Haijie and Wu, Chenming and Shi, Yahao and Cheng, Xinhua and Zhao, Chen and Feng, Haocheng and Ding, Errui and Wang, Jingdong and others},
  journal={arXiv preprint arXiv:2406.02058},
  year={2024}
}

@article{kerbl20233d,
  title={3d gaussian splatting for real-time radiance field rendering.},
  author={Kerbl, Bernhard and Kopanas, Georgios and Leimk{\"u}hler, Thomas and Drettakis, George},
  journal={ACM Trans. Graph.},
  volume={42},
  number={4},
  pages={139--1},
  year={2023}
}

@inproceedings{kirillov2023segment,
  title={Segment anything},
  author={Kirillov, Alexander and Mintun, Eric and Ravi, Nikhila and Mao, Hanzi and Rolland, Chloe and Gustafson, Laura and Xiao, Tete and Whitehead, Spencer and Berg, Alexander C and Lo, Wan-Yen and others},
  booktitle={Proceedings of the IEEE/CVF international conference on computer vision},
  pages={4015--4026},
  year={2023}
}

@article{sethian1996fast,
  title={A fast marching level set method for monotonically advancing fronts.},
  author={Sethian, James A},
  journal={proceedings of the National Academy of Sciences},
  volume={93},
  number={4},
  pages={1591--1595},
  year={1996}
}

@inproceedings{lindenberger2023lightglue,
  title={Lightglue: Local feature matching at light speed},
  author={Lindenberger, Philipp and Sarlin, Paul-Edouard and Pollefeys, Marc},
  booktitle={Proceedings of the IEEE/CVF international conference on computer vision},
  pages={17627--17638},
  year={2023}
}

@article{zou2023segment,
  title={Segment everything everywhere all at once},
  author={Zou, Xueyan and Yang, Jianwei and Zhang, Hao and Li, Feng and Li, Linjie and Wang, Jianfeng and Wang, Lijuan and Gao, Jianfeng and Lee, Yong Jae},
  journal={Advances in neural information processing systems},
  volume={36},
  pages={19769--19782},
  year={2023}
}

\end{document}